%% file: main.tex
\documentclass[letterpaper, 10 pt, conference]{ieeeconf}

\IEEEoverridecommandlockouts                              % This command is only needed if 
\let\labelindent\relax                  % resolves conflict between IEEE template and enumitem package

\usepackage{cite}
\usepackage{amsmath,amssymb,amsfonts}
\usepackage{algorithmic}
\usepackage{graphicx}
\usepackage{textcomp}
\usepackage{xcolor}

\usepackage{listings} % For code
\usepackage{booktabs} % For tables
\usepackage{url} % For urls
\usepackage{enumitem} % For RQ1, RQ2, RQ3
\usepackage{caption} % For introduction image
\usepackage{hyperref} % For autoref
\usepackage{flushend} % balance last page

\definecolor{acin_red}{RGB}{186, 18, 43}
\definecolor{acin_gray}{RGB}{176, 176, 176}
\definecolor{acin_yellow}{RGB}{252, 204, 71}
\definecolor{TU_blue}{RGB}{0, 102, 153}

\providecommand{\Graph}{\mathcal{G}}
\providecommand{\GraphwithPos}{\mathcal{G_P}}
\providecommand{\GraphwithPoswithEdges}{\mathcal{G_{P, E}}}

\providecommand{\GraphwithPoswithGeneratedEdges}{\mathcal{G_{P, \overline{E}}}}

\usepackage{tikz}
\usepackage{pgfplots}
\usetikzlibrary{pgfplots.groupplots}
\pgfplotsset{compat=1.9,height=0.3\textheight,legend cell align=left,tick scale binop=\times}
\pgfplotsset{grid style={loosely dotted,color=darkgray!30!gray,line width=0.6pt},tick style={black,thin}}
\pgfplotsset{every axis plot/.append style={line width=0.8pt}}
\usetikzlibrary{shapes, arrows, positioning, fit}

\def\eg{\textit{e.g.,}}
\def\ie{\textit{i.e.,}}

\usepackage{acro}
\input{acronyms}

\def\BibTeX{{\rm B\kern-.05em{\sc i\kern-.025em b}\kern-.08em
    T\kern-.1667em\lower.7ex\hbox{E}\kern-.125emX}}

\begin{document}
\bstctlcite{IEEEexample:BSTcontrol}

\title{\LARGE \bf Relational Scene Graphs for\\Object Grounding of Natural Language Commands}

%\author{Anonymous authors}
\author{Julia Kuhn, Francesco Verdoja, Tsvetomila Mihaylova, and  Ville Kyrki
\thanks{This work was supported by the Research Council of Finland (decision 354909). The authors acknowledge the use of the MIDAS infrastructure of Aalto School of Electrical Engineering.
J.\ Kuhn, F.\ Verdoja, and V.\ Kirki are with the School of Electrical Engineering, Aalto University, Espoo, Finland. T. Mihaylova is with the School of Science, Aalto University, Espoo, Finland. Email: \{firstname.lastname\}@aalto.fi
}}

\maketitle
\thispagestyle{empty}
\pagestyle{empty}

\input{Chapters/Abstract}

\input{Chapters/Introduction}
\input{Chapters/RelatedWorks}
\input{Chapters/ProblemFormulation}
\input{Chapters/Methods}
\input{Chapters/Experiments}

\input{Chapters/Conclusion}
\input{Chapters/Acknowledgments}

\bibliographystyle{IEEEtran}
\bibliography{bibliography}

\end{document}

%% file: acronyms.tex
\DeclareAcronym{llm}{
  short = LLM,
  long  = large language model,
  short-plural-form = LLMs,
  long-plural-form  = large language models
}

\DeclareAcronym{nlp}{
  short = NLP,
  long  = natural language processing
}

\DeclareAcronym{vlm}{
  short = VLM,
  long  = vision language model,
  short-plural-form = VLMs,
  long-plural-form  = vision language models
}

\DeclareAcronym{vla_3d}{
  short = VLA-3D,
  long  = Vision and Language-guided Action in 3D Scenes
}

\DeclareAcronym{react}{
  short = REACT,
  long  = Real-time Efficient Attribute Clustering and Transfer for Updatable 3D Scene Graph
}

\DeclareAcronym{gpt_3_5}{
  short = GPT-3.5,
  long  = Generative Pre-trained Transformer - 3.5
}

\DeclareAcronym{gpt_4o}{
  short = GPT-4o,
  long  = Generative Pre-trained Transformer - 4 omni
}

\DeclareAcronym{gpt_5}{
  short = GPT-5,
  long  = Generative Pre-trained Transformer - 5
}

\DeclareAcronym{ml}{
  short = ML,
  long  = Machine Learning
}

\DeclareAcronym{3dsg}{
  short = 3DSG,
  long  = 3D scene graph,
  short-plural-form = 3DSGs,
  long-plural-form  = 3D scene graphs
}

%% file: Chapters/Abstract.tex
\begin{abstract}

Robots are finding wider adoption in human environments, increasing the need for natural human-robot interaction.
However, understanding a natural language command requires the robot to infer the intended task and how to decompose it into executable actions, and to ground those actions in the robot's knowledge of the environment, including relevant objects, agents, and locations.
This challenge can be addressed by combining the capabilities of \acp{llm} to understand natural language with \acp{3dsg} for grounding inferred actions in a semantic representation of the environment.
However, many \acp{3dsg} lack explicit spatial relations between objects, even though humans often rely on these relations to describe an environment.
This paper investigates whether incorporating open- or closed-vocabulary spatial relations into \acp{3dsg} can improve the ability of \acp{llm} to interpret natural language commands.
To address this, we implement two pipelines using off-the-shelf models: an \ac{llm}-based pipeline for target object grounding from open-vocabulary language commands and a \acl{vlm}-based pipeline to add open-vocabulary spatial edges to \acp{3dsg} from images captured while mapping.
Finally, we evaluate two \acsp{llm} across 14 scenes using 905 natural language statements (786 procedurally-generated, 119 human-authored) to assess performance on the downstream task of target object grounding.
Our study demonstrates that explicit spatial relations improve the ability of \acp{llm} to ground objects, and while open-vocabulary relation generation with \aclp{vlm} proves feasible from robot-captured images, our analysis did not yield evidence favoring either open- or closed-vocabulary relations.

\end{abstract}

%% file: Chapters/Introduction.tex
\section{Introduction}
\acresetall

The inherent versatility of human language presents many challenges for robots to comprehend and execute natural language commands.
For example, a command such as ``bring me the small plate from the table'' requires considerable contextual knowledge.
To interpret this command correctly, it is necessary to determine which plate and which table the user is referring to.
This challenge of linking language commands to the physical world is known as \emph{grounding}, and while seemingly straightforward, it poses a significant challenge for robots.

To improve the abilities for command interpretation of robots, \ac{nlp} has gained considerable interest in robotics for several decades.
Traditionally, robot commands have been restricted to a closed vocabulary, which relies on a fixed set of predefined terms. 
Free-form natural language commands are instead open-vocabulary.
\Acp{llm} are well-suited for understanding open-vocabulary inputs, because they are trained on vast amounts of human-written text, and their capabilities have steadily increased over the past years. 
Consequently, their integration into robotics has shown considerable potential~\cite{huang2022languagemodelszeroshotplanners, ahn2022icanisay}.

However, grounding language in the environment goes beyond command interpretation and requires perceptual and reasoning capabilities that provide contextual knowledge of the objects and locations referenced in the scene.
This knowledge is often recorded in a structured representation, such as a map, tasked to provide both geometric and semantic information.
Recently, \acp{3dsg} have been proposed as an effective way to map and subsequently navigate an environment~\cite{hughes2022hydrarealtimespatialperception, nguyen2025reactrealtimeefficientattribute}.
\acp{3dsg} represent an environment as a graph, with nodes representing entities in the environment (\eg{} objects, rooms, and agents), and edges representing the semantic relations between them.
While most existing \acp{3dsg} do not record explicit spatial relations between objects (\eg{} ``the box \emph{under} the bed''), some recent works proposed to encode these spatial relations through edges~\cite{zhang2024vla3ddataset3dsemantic, 11128059, gu2023conceptgraphsopenvocabulary3dscene, rotondi2025fungraphfunctionalityaware3d, wang2025gaussiangraph3dgaussianbasedscene}, given that humans often rely on them to specify objects in an environment.
These approaches include both closed-vocabulary edges, some of which are inferred from bounding boxes, and open-vocabulary edges that are inferred from \acp{llm} or \acp{vlm}.
However, it is not known if these edges aid \acp{llm} in parsing \acp{3dsg} to ground objects.

\begin{figure}
    \centering
    \includegraphics[width=1\linewidth]{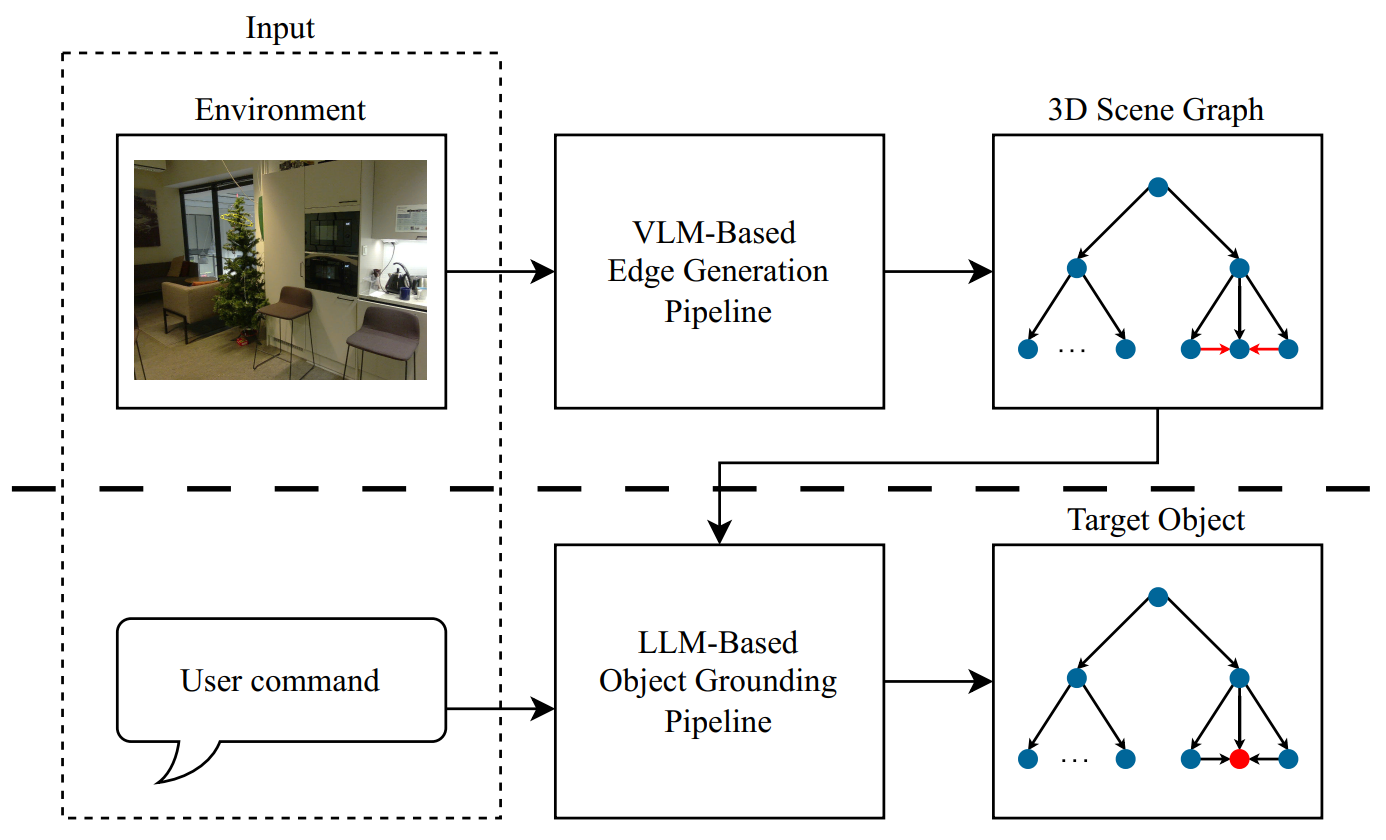}
    \caption{Overview of the setup. From images captured during mapping, spatial edges are added to a \acf{3dsg}. When a user command is issued, it is grounded through the \ac{3dsg}, leveraging these edges.}
    \label{fig:introduction_overview}
\end{figure}

In particular, two questions remain open, which we address in this work:
\begin{enumerate}[label=RQ\arabic*]
    \item Does incorporating spatial edges improve the ability of \acp{llm} to ground natural language commands?
    \item Considering that these edges are available, which approach to vocabulary yields better performance for the \acp{llm} to ground natural language commands: a closed or open-vocabulary one?
\end{enumerate}

To address the first research question, this paper evaluates the ability of two different \acp{llm} from OpenAI~\cite{openai_models2025}, specifically \ac{gpt_4o} and \ac{gpt_5}, to leverage spatial relations in \acp{3dsg} to understand natural language commands.
To address the second research question, we implement an open-vocabulary relation generation pipeline for \acp{3dsg} using an off-the-shelf \ac{vlm} to infer spatial relations, enabling comparison with the closed-vocabulary edges produced by~\cite{zhang2024vla3ddataset3dsemantic}.
In \autoref{fig:introduction_overview}, we provide an overview of the integration, where the top line represents the \ac{vlm}-based edge generation module and the bottom line the \ac{llm}-based target object grounding module.

The main contributions of this paper are:
\begin{itemize}
    \item Implementation of an \ac{llm}-based pipeline to ground target objects in open-vocabulary commands to an environment represented by a \ac{3dsg}.
    \item A method for extracting and labeling relevant images from the robot's mapping data, which are then used with a \ac{vlm}-based pipeline employing an off-the-shelf model to generate spatial edges for \acp{3dsg}.
    \item An extensive study of the performance of two different \acp{llm} tasked to ground target objects in open-vocabulary commands to an environment represented by different \ac{3dsg} variations.
\end{itemize}

%% file: Chapters/RelatedWorks.tex
\section{Related Works}

To operate in the real world, robots benefit from structured models of their environment.
There are various ways to represent the environment, and although the more traditional metric maps (\eg{} occupancy grid maps~\cite{1087316}) and semantic maps~\cite{app10020497} are widely adopted, \acp{3dsg} have found increasing adoption in robotics for their ability to efficiently represent both semantics and structure in a single compact abstraction.

\acp{3dsg} represent an environment as a graph.
Typically, \acp{3dsg} consist of nodes that represent entities, ranging from objects (with their distinct attributes) to rooms or agents.
The nodes are connected through edges that represent the semantic relations between them.

One widely used \ac{3dsg} implementation is Hydra~\cite{hughes2022hydrarealtimespatialperception}, which enables real-time scene mapping and subsequent robotic navigation.
\ac{react}~\cite{nguyen2025reactrealtimeefficientattribute} extends Hydra by enabling autonomous robots to track and update object states over time.
While those \acp{3dsg} offer a solid structural representation of an environment, object-object relations are missing, even though humans often describe objects through them.

In contrast, approaches such as ConceptGraphs~\cite{gu2023conceptgraphsopenvocabulary3dscene} explicitly represent spatial object-object relations.
Spatial relations describe the relative position of two or more entities in 3D space, such as ``lamp \emph{above} bed'' or ``suitcase \emph{between} chair and bed''.
One approach to extracting spatial relations is to analyze the bounding boxes of detected entities to infer geometric relations, as demonstrated in the \ac{vla_3d} dataset~\cite{zhang2024vla3ddataset3dsemantic}.
That dataset primarily aims to create a large and diverse collection of indoor scenes, focusing on extracting \ac{3dsg} edges and navigation commands in natural language.

Some relations cannot be inferred solely from the objects and their positions.
To extract relations that require a deeper understanding of interactions, a growing focus has been on visual relationship detection (VRD) as a research area~\cite{9693341}.

Relations can be represented using either closed-vocabulary approaches \cite{zhang2024vla3ddataset3dsemantic, 11128059} or open-vocabulary approaches \cite{gu2023conceptgraphsopenvocabulary3dscene, rotondi2025fungraphfunctionalityaware3d, wang2025gaussiangraph3dgaussianbasedscene} derived from \acp{llm} or \acp{vlm}.
Closed-vocabulary relies on a fixed set of predefined terms for each relation type, which simplifies their creation through defined rules, \eg{} with bounding boxes.
Additionally, using closed-vocabulary relations simplifies the interpretation of the relations.
In contrast, open-vocabulary relations instead adhere to natural language by not restricting the choice of words, intuitively enabling more precise and flexible descriptions of the relations between entities.

However, it remains an open question whether explicit spatial edges between objects, closed or open-vocabulary, help \acp{llm} on downstream tasks, which this paper investigates.
Specifically, for target object grounding, we evaluate geometry-derived, closed-vocabulary relations in \ac{vla_3d} and extend \ac{react} \acp{3dsg} with geometry-derived, closed-vocabulary edges as well as open-vocabulary spatial edges inferred from robot-captured images to compare their impact.
This positions our work as a downstream evaluation that links prior graph construction methods to their impact on language grounding.

%% file: Chapters/ProblemFormulation.tex
\section{Problem Formulation}

We define target object grounding as the task of resolving a natural language referential statement to a unique object in its environment.

A referential statement is a textual description that unambiguously identifies an object in a scene, \eg{}, ``the cabinet below the office desk'' refers to a specific cabinet.

Following the definition in \cite{zhang2024vla3ddataset3dsemantic}, we assume the environment to be represented as a \ac{3dsg}, \ie{} a graph with nodes that represent objects, including their names and IDs, as well as attributes such as colors and spatial positions in the form of bounding boxes.
The relations between the objects are represented by the edges of the graph, containing only spatial edges.
These edges are expressed either in closed-vocabulary, \eg{} \textit{on}, or in open-vocabulary, \ie{} natural language.
Both are treated as alternative representations of the same underlying spatial relation.
The closed-vocabulary edges are defined unidirectionally from one object to another.
An exception to this is the edge \textit{between}, which involves an object positioned in the middle, with the edges pointing unidirectionally to the two associated objects.

Given a referential statement and a \ac{3dsg}, the task is to correctly identify the referenced object by its ID.

%% file: Chapters/Methods.tex
\section{Grounding Natural Language Commands with \acsp{3dsg}}

The following sections present our setup, which is divided into two parts: grounding natural language commands with the structured \ac{3dsg} and enhancing the \ac{3dsg}'s expressiveness by refining relations between objects to capture spatial descriptions commonly used in natural language.

\subsection{Enabling \acsp{llm} to Reason over \acsp{3dsg}}

\subsubsection{Serializing \acsp{3dsg} for \acsp{llm}}
\label{subsubsec:Preprocessing Scene Graphs for LLM Input}

A raw \ac{3dsg} cannot be interpreted by an \ac{llm} and must be converted into a textual representation that includes only the necessary information to interpret the scene.
To facilitate the \ac{llm}'s processing of the information, both nodes and edges need to be serialized into text.

First, for each node, the object's name, the unique identifier, and its attributes are converted into string format.
Explicitly naming the objects can add semantic context to the scene since the \ac{llm} may possess knowledge about particular objects or relations between them from its training data.
For instance, an \ac{llm} understands that a mug is a small object and that the likelihood of the mug being on the table is greater than it being placed on a chair situated next to the table.
The unique identifier (ID) is necessary to distinguish between multiple instances of the same object type.
Attributes such as color and size can enhance the \ac{llm}'s understanding of the scene, allowing for better differentiation between objects of the same class.
Additionally, position information contributes to a general comprehension of the scene.

Second, to make the spatial edges understandable for the \ac{llm}, the unique identifiers for the objects along with the corresponding relation between them are used.
For the edges in this implementation, the object's name or other attributes are not considered because they are already present in the nodes and do not provide enough valuable information to warrant the extra space.

\subsubsection{Object Grounding with \acsp{llm}}
\label{subsubsec:Pipeline for LLM}

\begin{figure}[t!]
    \centering
    \includegraphics[width=1\linewidth]{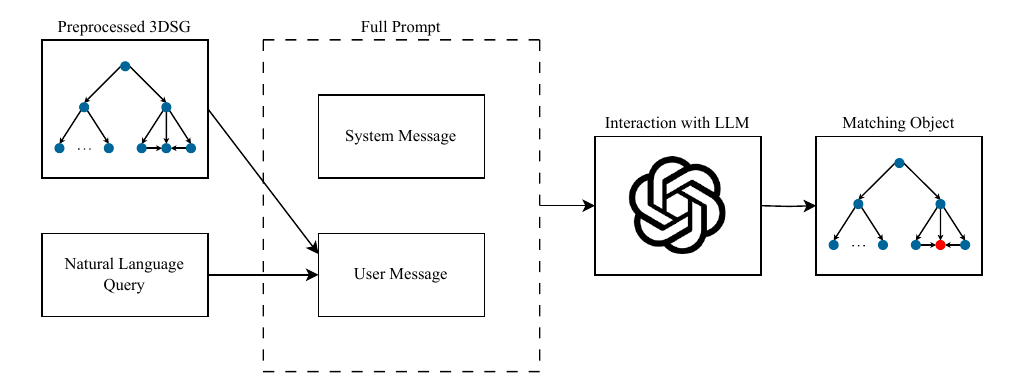}
    \caption{Overview of the \acs{llm}-based pipeline for object grounding in \acsp{3dsg}.}
    \label{fig:pipeline_llm}
\end{figure}

To evaluate grounding performance, we enable direct querying of the serialized \ac{3dsg} by an \ac{llm}.
To address this, a pipeline is implemented that enables the \ac{llm} to query the serialized \ac{3dsg} directly, facilitating the grounding of objects in the scene based on natural language commands.
An overview of the pipeline structure is shown in \autoref{fig:pipeline_llm}.

The objective of the pipeline is to query for a particular object in a \ac{3dsg} based on a natural language referential statement.
It takes two inputs: the processed \ac{3dsg} and the natural language command.
The two inputs are sent via a prompt\footnote{All prompts used are available at \url{https://github.com/aalto-intelligent-robotics/object_grounding_scene_graph}}, which comprises two parts: a system message that defines the behavior of the \ac{llm} on a higher level and a user message that contains both the serialized \ac{3dsg} and the natural language command.
The expected \ac{llm} output is the ID of the referenced object that best responds to the query.
To avoid biased outputs, prompts are sent without a chat memory.
Thus, both the system and the user message are included with every request.

\subsection{Generating Spatial Relations from Visual Data}

% We implement a \ac{vlm}-based pipeline to extract relations expressed in natural language between objects from images.
% We assume that this approach could enhance the \ac{3dsg}'s expressiveness by using real-world knowledge of the \ac{vlm} to integrate richer semantic information into the relations.

While bounding boxes suffice for rule-based, closed-vocabulary relations, they do not capture the open-vocabulary descriptions used in natural language.
We therefore use an off-the-shelf \ac{vlm} to obtain open-vocabulary spatial relations from images, leveraging the model’s learned real‑world knowledge.

\subsubsection{Preparing Scene Images for \ac{vlm}-based pipeline}
\label{subsubsec:Preparing Scene Images for Relation Extraction}

\begin{figure}[t!]
    \centering
    \includegraphics[width=0.8\linewidth]{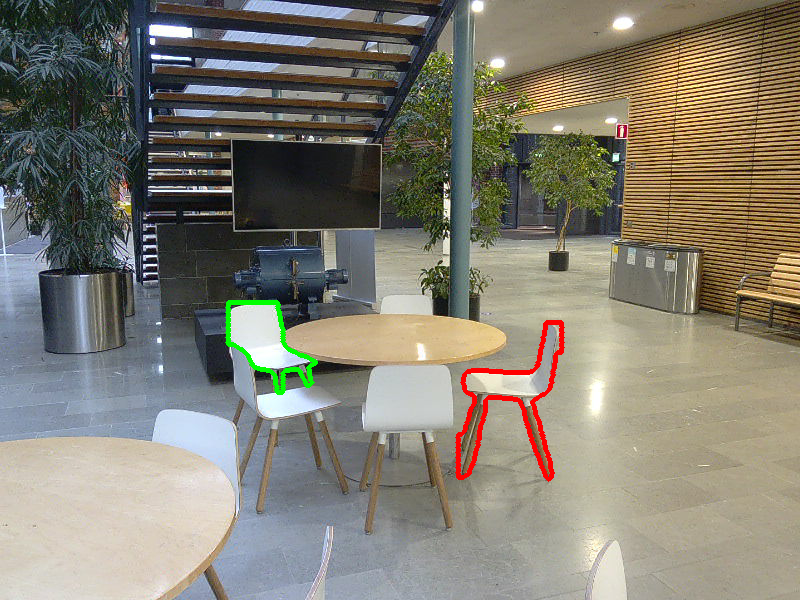}
    \caption{Example image used for edge generation, where the generated edge is: \textit{both are positioned around the same table, facing each other}}
    \label{fig:ex2_edge_generation_1}
\end{figure}

To enable the \ac{vlm} to focus on extracting spatial relations between two objects rather than locating them within the image, a preprocessing step for the images is required.
This paper proposes to pre-mark the objects of interest by outlining them in color, ensuring their positions are easily visible to the \ac{vlm} (\eg{} \autoref{fig:ex2_edge_generation_1}).
The outlining is performed by overlaying segmentation masks from a segmentation model onto the original image.

Not all \acp{3dsg} contain images of the scenes, which limits this approach to those that do, \eg{} \acs{react}~\cite{nguyen2025reactrealtimeefficientattribute}.
Additionally, it is crucial to identify which exact object is present in each image to generate valid relations.

\subsubsection{Inferring Spatial Relations with \acsp{vlm}}
\label{subsubsec:Inferring Spatial Relations with VLMs}

\begin{figure}[t!]
    \centering
    \includegraphics[width=1\linewidth]{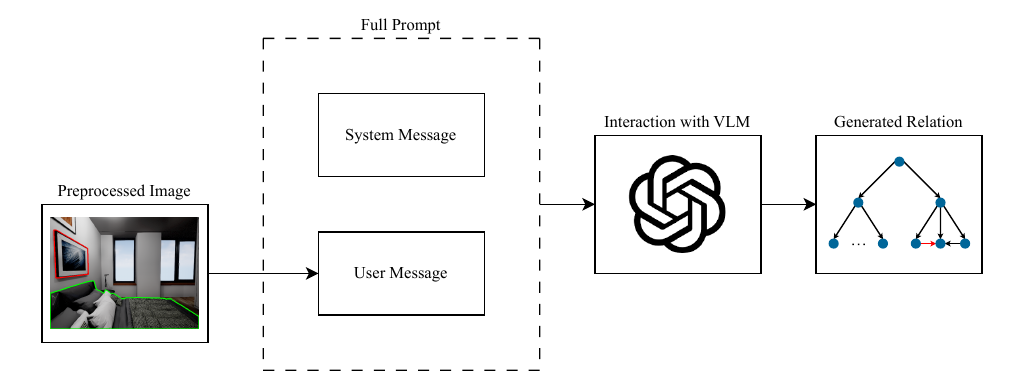}
    \caption{Overview of the \acs{vlm}-based pipeline for relation generation.}
    \label{fig:pipeline_vlm}
\end{figure}

The pipeline is used to generate spatial relations between objects in the scene by utilizing the preprocessed images.
An overview of the pipeline is illustrated in \autoref{fig:pipeline_vlm}.

The system message remains constant across API requests, whereas the user message varies by incorporating an image that specifies the relation to be generated.
The output of the \ac{vlm} consists of the generated relation between the objects in open-vocabulary format.
To ensure consistency in format and avoid excessively long edge descriptions, the \ac{vlm} is instructed to structure and shorten its output.
An output example is given in the caption of \autoref{fig:ex2_edge_generation_1}.

%% file: Chapters/Experiments.tex
\section{Experiments}

The experiments were designed to investigate whether \acp{llm} can leverage spatial edges in \acp{3dsg}.
Specifically, Experiment 1 aimed to address RQ1 on the downstream task of target object grounding, while Experiment 2
evaluated whether open-vocabulary edges generated from images impact object grounding performance differently from closed-vocabulary ones (RQ2).
We conducted our experiments across two datasets: the \ac{vla_3d} dataset~\cite{zhang2024vla3ddataset3dsemantic} and the \ac{react} dataset~\cite{nguyen2025reactrealtimeefficientattribute}.

\subsection{Evaluation Procedure}
\label{subsec:Evaluation}

Given a \ac{3dsg} and a natural language command, the goal for the \ac{llm} was to correctly identify the referenced object in the referential statement.
The \ac{llm} had to return the ID of the object referenced in the referential statement.
To evaluate the \ac{llm} performance on this task, the accuracy was calculated by dividing the number of correct predictions by the total number of predictions.
A prediction was considered correct when the output ID matched the ground truth ID.
Predictions where the \ac{llm} output was in an invalid format were counted as incorrect.

To determine whether performance differences between two methods were statistically significant, McNemar's test \cite{mcnemar1947sampling} was employed.
This test is well-suited for analyzing paired and binary data.

\subsection{Experimental Setup}
\label{sec:Experimental Setup}

\subsubsection{Description of \acs{vla_3d} Dataset}
\label{subsec:Description of VLA-3D Dataset}

The \ac{vla_3d} dataset~\cite{zhang2024vla3ddataset3dsemantic} contains 7635 indoor scenes obtained from real-world environments~\cite{baruch2022arkitscenesdiverserealworlddataset, ramakrishnan2021habitatmatterport3ddatasethm3d, chang2017matterport3dlearningrgbddata, dai2017scannetrichlyannotated3dreconstructions, wald2019rio3dobjectinstance} and scenes synthetically generated with Unity~\cite{unity}. 
Each scene contains between 4 and 2264 objects and can be composed of multiple rooms.
For each scene, 3D point clouds and object metadata, including bounding boxes, positions, and colors are provided, together with a semantic \acp{3dsg} generated using the procedure described in the \ac{vla_3d} paper~\cite{VLA3D_GitHub}.
These \acp{3dsg} include objects as nodes and their spatial relations as labeled edges.
The relation types fall into three categories: binary (\eg{} above), ordered (\eg{} closest), and ternary (\eg{} between).
The bounding box information serves as the basis for deriving the relation between the objects.
Moreover, for each scene, a list of textual descriptions is provided.
Each of these descriptions references a target object and describes it in terms of a specific spatial edge in the scene's \ac{3dsg} connecting it to another object.
To prevent ambiguous referential statements, the generator appends color or size information when necessary.
Although the generated referential statements are semantically rich, the reliance on sentence templates and the use of synonyms limits the linguistic variety.

\subsubsection{Preprocessing of \acs{vla_3d} Dataset}
\label{subsec:Preprocessing of VLA-3D Dataset}

The dataset was preprocessed to provide the relevant information in a compact format for the \ac{llm} input.
From the \acp{3dsg}, we extracted the following attributes for each object: \texttt{object\_id}, \texttt{nyu\_label}, \texttt{color\_labels}, \texttt{center}, and \texttt{size}.
The extracted data allowed the processing of each \ac{3dsg} into a compact format suitable for the model, as described in \autoref{subsubsec:Preprocessing Scene Graphs for LLM Input}.
This resulted in the formatted edge \texttt{obj1\_id|relation|obj2\_id} for binary and ordered relation types, where \texttt{obj1\_id} represents the ID of the target object and \texttt{obj2\_id} represents the ID of the anchor.
For the ternary relation type \textit{between}, the target object is between two anchor objects with the ids \texttt{obj2\_id} and \texttt{obj3\_id}, which resulted in the formatted edge \texttt{obj1\_id|relation|obj2\_id and obj3\_id}.

\iffalse
\begin{table}[t!]
    \centering
    \caption{Overview from the used subset of the \ac{vla_3d} dataset.}
    \begin{tabular}{llcc}
        \toprule
        \textbf{Dataset} & \textbf{Scene Name}                   & \textbf{Objects} & \textbf{Referential} \\
                         &                                       &                  & \textbf{Statements} \\
        \midrule
        
        3RScan           & 0a4b8ef6-a83a-21f2-8672-dce34dd0d7ca  & 30               & 25  \\
                         & 0ad2d3a1-79e2-2212-9b99-a96495d9f7fe  & 21               & 9   \\
                         & 0ad2d3a3-79e2-2212-9a51-9094be707ec2  & 18               & 12  \\
                         & 0ad2d38f-79e2-2212-98d2-9b5060e5e9b5  & 25               & 11  \\
                         & 0ad2d39b-79e2-2212-99ae-830c292cd079  & 30               & 30  \\
                         & 0ad2d39d-79e2-2212-99aa-1e95c5b94289  & 28               & 23  \\
        \midrule
        
        Scannet          & scene0000\_00                         & 67               & 163 \\
                         & scene0000\_01                         & 72               & 231 \\       
        \midrule
        
        Unity            & studio                                & ?                & 208 \\
        \bottomrule
    \end{tabular}
    \label{tab:subset_vla}
\end{table}
\fi

\begin{table}[t!]
    \centering
    \caption{Overview from the used subset of the \ac{vla_3d} dataset.}
    \begin{tabular}{llc}
        \toprule
        \textbf{Dataset} & \textbf{Scene Name}                   & \textbf{Statements} \\
        \midrule
        
        3RScan           & 0a4b8ef6-a83a-21f2-8672-dce34dd0d7ca  & 25  \\
                         & 0ad2d3a1-79e2-2212-9b99-a96495d9f7fe  & 9   \\
                         & 0ad2d3a3-79e2-2212-9a51-9094be707ec2  & 12  \\
                         & 0ad2d38f-79e2-2212-98d2-9b5060e5e9b5  & 11  \\
                         & 0ad2d39b-79e2-2212-99ae-830c292cd079  & 30  \\
                         & 0ad2d39d-79e2-2212-99aa-1e95c5b94289  & 23  \\
        \midrule
        
        Scannet          & scene0000\_00                         & 163 \\
                         & scene0000\_01                         & 231 \\       
        \midrule
        
        Unity            & studio                                & 208 \\
        \midrule
        
        \textbf{Total}   &                                       & \textbf{712} \\
        \bottomrule
    \end{tabular}
    \label{tab:subset_vla}
\end{table}

A subset of the \ac{vla_3d} dataset listed in \autoref{tab:subset_vla} was used in the evaluation.
It comprised nine scenes from three different sources~\cite{wald2019rio3dobjectinstance, dai2017scannetrichlyannotated3dreconstructions, unity}, chosen randomly among real-world and synthetic scenes consisting of a single room.
The relations \textit{closest} and \textit{farthest} were excluded from the edges and the referential statements.
These relations were defined for every possible pair of objects, creating excessive and often irrelevant relations.
To test for a more generalized understanding, we sampled the available referential statements.
When multiple synonymous referential statements were available for a single underlying relation, \eg{} \textit{below} has synonyms \textit{under} and \textit{beneath}, only one of these statements was randomly selected.
These processing steps ensured that only non-redundant and varied referential statements were retained.

\subsubsection{Description of \acs{react} Dataset}

The \ac{react} dataset~\cite{nguyen2025reactrealtimeefficientattribute} comprises three different real-world indoor scenes captured with Hello Robot Stretch 2~\cite{kemp2022designstretchcompactlightweight}.
These scenes are annotated according to the COCO~\cite{lin2015microsoftcococommonobjects} label space, but only for the three most prominent object categories of the scene: chair, couch, and dining table.
Furthermore, it contains two synthetically made indoor scenes obtained from the FLAT dataset~\cite{Schmid_2022}.
In contrast, these scenes had their own label space, with the following object categories: bed, chair, coffee table, floor lamp, journal, picture, sofa, and table.

\subsubsection{Preprocessing of \acs{react}}

First, the raw data from the real-world scenes were again processed using the \ac{react} framework to enrich the annotations.
The \ac{react} pipeline uses YOLOv11~\cite{khanam2024yolov11overviewkeyarchitectural}, pre-trained on the COCO dataset, for object detection and instance segmentation.
By using the whole COCO label space instead of the three former categories, the number of categories increased to ten.
This increased the number and diversity of the relations between the objects. 

The \acp{3dsg} produced by the \ac{react} pipeline over the dataset were converted into the \ac{vla_3d} format and then passed through the same preprocessing described in \autoref{subsubsec:Preprocessing Scene Graphs for LLM Input} to convert the \acp{3dsg} into a format usable by the \ac{llm}.
Moreover, relational edges and the corresponding referential statements were created for this dataset using the \ac{vla_3d} codebase, resulting in 74 referential statements over the five scenes of the dataset.

To generate edges between objects from visual data, images with corresponding highlighted objects were needed, as described in \autoref{subsubsec:Preparing Scene Images for Relation Extraction}.
For each detected object, the \ac{react} framework saved the set of images in which the object appeared, along with its corresponding segmentation masks.
To find the image in which the two given objects are most visible, all pictures containing both objects were examined.
The pixel count for their corresponding segmentation masks was computed, selecting the picture with the highest combined pixel count.
Although this approach could fail when object sizes differ greatly, it worked reliably for this dataset.

\subsubsection{Creation of \ac{react} Dataset with Human Commands}

In order to extend our evaluation beyond the templated referential statements generated following the \ac{vla_3d} approach, we collected statements from human annotators.

For every object in the \ac{react} scenes, the image containing the highest number of pixels belonging to that object was selected and outlined in color.
This resulted in 104 images with outlined objects, which were split into 10 batches of 10 or 11 images, and each batch was sent to two different annotators.
The following instruction was given to the annotators: ``Your friend is in the room shown in the photo. Describe the highlighted object so they can recognize or locate it''.
For each object, two descriptions were obtained, resulting in a total of 208 referential statements.
The descriptions were then evaluated by the authors for ambiguity.
All ambiguous descriptions (\ie{} more than one object in the image matched the descriptions) were discarded, resulting in a final set of 119 statements.
We refer to this final set of statements as the \textit{\ac{react} human command dataset}.

\subsubsection{Models and Prompt}
\label{subsec:Models}

In Experiment 1, the models \ac{gpt_4o} and \ac{gpt_5} from OpenAI~\cite{openai_models2025} were used.
The system message included an explanation of the task, input and expected output values, a definition of all spatial edges, and an example.
The expected output was the ID of the target object specified in the referential statement, which was incorporated into the user message of the prompt.
The serialized \ac{3dsg} was inserted in the user message.
For the model \ac{gpt_5}, the system and user messages were combined into a single message, because the model expects only one input.

In Experiment 2, \ac{gpt_4o} was prompted to generate open-vocabulary spatial edges from images.
The system message in the prompt included an explanation of the task and the expected output format.
The preprocessed image was inserted into the user message, along with the names and IDs of the two objects whose relation was being examined.

\subsection[Experiment 1]{Experiment 1: Impact of Spatial Edges on \acs{llm} Scene Understanding}
\label{subsec:Experiment 1}

Experiment 1 was conducted to analyze the difference in \ac{llm} scene understanding between a \ac{3dsg} with and without spatial edges.
Therefore, it investigated whether \acp{llm} can connect positional information in the form of the objects' bounding boxes with spatial edges for improved target object grounding.

\subsubsection{Experiment design}
\label{subsec:Ex1: Experiment design}

To evaluate the impact of spatial edges, three different variants for input graphs were defined: $\Graph$, $\GraphwithPos$, and $\GraphwithPoswithEdges$.
$\Graph$ served as the baseline; it contained only nodes with the object's name and ID, and no edges.
$\GraphwithPos$ added color and a bounding box to each node, and $\GraphwithPoswithEdges$ added the edges of the \ac{3dsg}, which represent the relations between the objects.

Because the baseline graph $\Graph$ contained only the object names, when an object was referenced by name, there was an equal chance for any of the objects with the same name to be retrieved.
Therefore, the baseline graph $\Graph$ was evaluated with the following calculation:
For each object class, a set $\mathcal{O}_C$ was created that contained all objects of that specific class.
Since objects of the same class could not be distinguished by name alone, the accuracy of each object was calculated by using the probability of randomly selecting the target object from its set $\mathcal{O}_C$, \ie{} $\frac{1}{|\mathcal{O}_C|}$.
For the statistical test, one object was randomly selected from $\mathcal{O}_C$.

The graphs $\GraphwithPos$ and $\GraphwithPoswithEdges$ were both evaluated separately with the pipeline described in \autoref{subsubsec:Pipeline for LLM}.

\subsubsection{Results}

In \autoref{tab:accuracies_total}, the accuracy of the target object grounding task is shown.
For \ac{gpt_4o} on the \ac{vla_3d} dataset, the accuracy improved by 3.97\% from $\Graph$ to $\GraphwithPos$ and by 7.3\% from $\GraphwithPos$ to $\GraphwithPoswithEdges$, achieving an accuracy of 84.27\%.
The accuracy on the \ac{react} dataset with \ac{gpt_4o} increased from the baseline $\Graph$ to the input graph $\GraphwithPos$ by 8.43\%.
From $\GraphwithPos$ to $\GraphwithPoswithEdges$, an increase of the accuracy of 9.46\% can be seen.
For \ac{gpt_5} on the \ac{vla_3d} dataset, $\GraphwithPoswithEdges$ outperformed $\GraphwithPos$ by 1.41\% and $\Graph$ by 26.58\% with a total accuracy of 99.58\%.
On the \ac{react} dataset, the accuracy was more than doubled when comparing the baseline $\Graph$ to $\GraphwithPos$, and 48.97\% higher for $\GraphwithPoswithEdges$ compared to $\Graph$.
For both datasets, edges had a considerable effect on the performance of both \acp{llm}, with \ac{gpt_5} outperforming \ac{gpt_4o} in all cases.

\autoref{tab:mcnemar_total} presents the results of McNemar's test for \ac{gpt_4o} and \ac{gpt_5}.
For \ac{gpt_4o}, no statistically significant differences were observed when comparing the baseline $\Graph$ with the graph $\GraphwithPos$.
When comparing $\Graph$ with $\GraphwithPoswithEdges$ on both datasets, a statistical significance was shown, from $\GraphwithPos$ to $\GraphwithPoswithEdges$ only on the \ac{vla_3d} dataset.
For \ac{gpt_5}, all differences were statistically significant.

\begin{table}[t!]
    \centering
    \caption{Total accuracy of the target object grounding task for $\Graph$, $\GraphwithPos$, and $\GraphwithPoswithEdges$ on the \acs{vla_3d} and \acs{react} datasets.}
    \begin{tabular}{llll}
        \toprule
                     & \textbf{Graph}            & \textbf{\ac{vla_3d}}         & \textbf{\ac{react}} \\
        \midrule
        
        Random       & $\Graph$                  & 0.7300                       & 0.3617  \\
        \midrule
        
        % \ac{gpt_3_5} & $\GraphwithPos$           & 0.6952                     & -     \\
        %              & $\GraphwithPoswithEdges$  & 0.632                      & -     \\      
        % \midrule
        
        \ac{gpt_4o}  & $\GraphwithPos$           & 0.7697                       & 0.4459 \\
                     & $\GraphwithPoswithEdges$  & 0.8427                       & 0.5405 \\       
        \midrule
        
        \ac{gpt_5}  & $\GraphwithPos$           & 0.9817                        & 0.7297 \\
                    & $\GraphwithPoswithEdges$  & \textbf{0.9958}               & \textbf{0.8514} \\
        \bottomrule
    \end{tabular}
    \label{tab:accuracies_total}
\end{table}

\begin{table}[t!]
    \centering
    \caption{McNemar's test results for Experiment 1. The significance level was categorized as follows: * significant ($p < 0.05$), ** highly significant ($p < 0.01$), and *** very highly significant ($p < 0.001$).}
    \begin{tabular}{llll}
        \toprule
                    & \textbf{Comparison}                         & \textbf{\ac{vla_3d}} & \textbf{\ac{react}} \\
        \midrule

        \ac{gpt_4o} & $\Graph$ vs $\GraphwithPos$                 &                      &    \\
                    & $\Graph$  vs $\GraphwithPoswithEdges$       & ***                  & ** \\
                    & $\GraphwithPos$ vs $\GraphwithPoswithEdges$ & ***                  &    \\
        \midrule
        
        \ac{gpt_5}  & $\Graph$  vs $\GraphwithPos$                & ***                  & *** \\
                    & $\Graph$  vs $\GraphwithPoswithEdges$       & ***                  & *** \\
                    & $\GraphwithPos$ vs $\GraphwithPoswithEdges$ & **                   & *   \\
        \bottomrule
    \end{tabular}
    \label{tab:mcnemar_total}
\end{table}

\subsection[Experiment 2]{Experiment 2: Influence of Open-Vocabulary Edges}
\label{subsec:Experiment 2}

In this experiment, the question of whether open-vocabulary spatial edges provide benefits for the \ac{llm}-based object grounding compared to the closed-vocabulary ones was investigated.

\subsubsection{Experiment Design}

For this experiment, the \ac{vlm} only generated edges that are also present in the graph $\GraphwithPoswithEdges$.
This enables using the same referential statements for open and closed-vocabulary edges.
If an edge was present in the dataset but could not be estimated because there was no fitting image, the original closed-vocabulary edge was used instead.
This resulted in the input graph $\GraphwithPoswithGeneratedEdges$.
Only reference statements with associated edges that were newly generated were tested, which resulted in 26 reference statements.
Of these, 22 had the edge \textit{near}, and 4 had the edge \textit{on}.
Furthermore, the human-generated commands of the \ac{react} dataset were evaluated.

The same setup as in Experiment 1 was then used with $\GraphwithPoswithGeneratedEdges$ and compared against the results with the input graph $\GraphwithPoswithEdges$.

\subsubsection{Results}

The accuracy on the \ac{react} dataset with generated commands and \ac{react} human command dataset of the graph with the closed-vocabulary edges $\GraphwithPoswithEdges$ and the generated open-vocabulary edges $\GraphwithPoswithGeneratedEdges$ is shown in \autoref{tab:ex2_accuracies_total}.

\begin{table}[t!]
    \centering
    \caption{Total accuracy of the target object grounding task for $\GraphwithPoswithEdges$ and $\GraphwithPoswithGeneratedEdges$ on the \acs{react} dataset.}
    \begin{tabular}{llcc}
        \toprule
                    & \textbf{Graph}                    & \textbf{Generated Commands}  & \textbf{Human Commands} \\
        \midrule
        
        \ac{gpt_4o} & $\GraphwithPoswithEdges$          & 0.7308                      & 0.4202 \\
                    & $\GraphwithPoswithGeneratedEdges$ & 0.6923                      & 0.3613 \\
        \midrule
        \ac{gpt_5}  & $\GraphwithPoswithEdges$          & 0.8077                      & \textbf{0.6218} \\
                    & $\GraphwithPoswithGeneratedEdges$ & \textbf{0.8462}             & 0.5882 \\
        \bottomrule
    \end{tabular}
    \label{tab:ex2_accuracies_total}
\end{table}

On the \ac{react} dataset with the generated commands, $\GraphwithPoswithGeneratedEdges$ achieved 3.85\% higher accuracy compared to $\GraphwithPoswithEdges$ for \ac{gpt_5}.
In contrast, for \ac{gpt_4o}, the accuracy for $\GraphwithPoswithGeneratedEdges$ was 3.7\% lower than for $\GraphwithPoswithEdges$.
On the \ac{react} human command dataset, \ac{gpt_4o} yielded an accuracy decrease of 5.89\% and \ac{gpt_5} resulted in a 3.36\% decrease.
These differences in accuracy between the closed-vocabulary edges and the generated open-vocabulary edges were not statistically significant.

\begin{figure}[t!]
    \centering
    \includegraphics[width=0.8\linewidth]{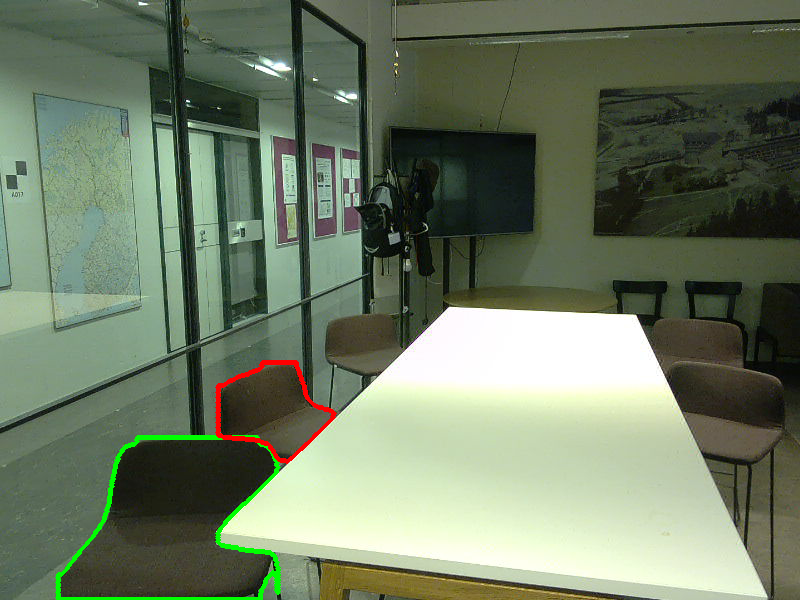}
    \caption{Example image used for edge generation, where the generated edge is: \textit{left side of the table, closer to the camera}}
    \label{fig:ex2_edge_generation_2}
\end{figure}

For all of the closed-vocabulary edges with the relation \textit{on}, the open-vocabulary version chosen by the \ac{vlm} was \textit{on top of}.
Several different edges were generated instead of the closed-vocabulary relation \textit{near}.
For instance, in \autoref{fig:ex2_edge_generation_1} and \autoref{fig:ex2_edge_generation_2}, a similar setting was depicted, where two chairs were positioned around the same table.
The resulting edges are: \textit{both are positioned around the same table, facing each other} and \textit{left side of the table, closer to the camera}.
Other examples for the generated edge from the closed-vocabulary relation \textit{near} included \textit{closer to the glass wall and stairs than} and \textit{near center, in front of a long white table, red chair is far to the back right corner}.

\subsection{Discussion}

The results of Experiment 1 revealed how two different \acp{llm} leverage \acp{3dsg} for target object grounding.
\ac{gpt_4o} and, especially \ac{gpt_5} consistently benefited from the inclusion of spatial edges on both datasets.

For \ac{gpt_4o} and \ac{gpt_5} the performance steadily improved from $\Graph$ to $\GraphwithPos$, as well as from $\GraphwithPos$ to $\GraphwithPoswithEdges$.
This suggests that these models are capable of reasoning with positional inputs and that spatial edges are beneficial for these \acp{llm} to understand object relations.
Overall, \ac{gpt_5} demonstrated the best accuracy in every category, reaching an impressive 99.5\% accuracy on \ac{vla_3d} and 85\% on \ac{react}.

As a note, the model \ac{gpt_3_5} was also tested on the \ac{vla_3d} dataset, but its accuracy declined as more information was given.
This implies that the model struggles to reason with additional information and is even hindered by it.
This indicates that for older \acp{llm}, it may be beneficial to reduce the additional information to achieve better performance.

Of the two datasets, the \ac{react} dataset proved to be more challenging, with a decrease in accuracy of 30.22\% for \ac{gpt_4o} and 14.44\% for \ac{gpt_5} for $\GraphwithPoswithEdges$ when compared to the \ac{vla_3d} dataset.
This may be attributed to the more challenging environment, where many objects belonged to the same class and often shared similar attributes (see \autoref{fig:ex2_edge_generation_1} and \autoref{fig:ex2_edge_generation_2}), resulting in ambiguous edges.

%%%%%%%%%%%%% DISCUSSION EXPERIMENT 2 %%%%%%%%%%%%%

The differences in accuracy between the closed-vocabulary edges and the generated edges from Experiment 2 were not statistically significant, indicating that open-vocabulary relations generated from images neither consistently outperform nor significantly underperform closed-vocabulary relations for the object grounding task.
Due to the small scale of the \ac{react} dataset, the lack of statistical significance may partly stem from insufficient data rather than an actual absence of performance difference.

There were both successful and unsuccessful examples of the generated edges.
A successful example included the edge \textit{both are positioned around the same table, facing each other}, which resulted in a more precise and semantically meaningful relation.
In contrast, an unsuccessful example was the edge \textit{left side of the table, closer to the camera}.
This edge was misleading, as the \ac{vlm} attempted to describe the scene related to the camera position at the time of capture.
However, the \ac{3dsg} contains no information about the camera or its positions while capturing the scene.
Additionally, when viewed from the opposite side of the table, the chairs are positioned to the right of the table, and the chair marked in red would be closer to the observer.
This failure suggests that the model struggles to generate view-independent edges.

When no straightforward relation existed between the objects, the model encountered difficulties.
As was suggested in the prompt, the model can use other objects to describe the relations between items.
For instance, it generated the edge \textit{closer to the glass wall and stairs than}, yet these objects are not part of the \ac{3dsg}, making it unclear how these edges compare qualitatively to the closed ones.
In some cases, the model generated edges that included the outline colors, \eg, \textit{near center, in front of a long white table, red chair is far to the back right corner}.
This presents a problem, as the \ac{3dsg} does not contain a \textit{red chair}.

Misleading or incorrectly generated edges can create potential issues, and filtering out these edges can be difficult.
The absence of a clear performance gap between the two edge representations suggests that the potential benefits of open-vocabulary relations may currently be limited less by their expressiveness and more by the reliability of their generation.

%% file: Chapters/Conclusion.tex
\section{Conclusion}

In this paper, we investigated how well \acp{llm} can leverage edges describing object relations in \acp{3dsg} to improve grounding of target objects, and whether open- or closed-vocabulary yields superior performance.

We demonstrated that incorporating explicit spatial information, including both positional information and spatial edges, can improve the performance of \acp{llm} for target object grounding in indoor environments.
Furthermore, our results did not indicate a statistically significant difference in performance when using closed or open-vocabulary edges in \acp{3dsg} for object grounding.

Although no statistically significant performance difference was observed, the result does not rule out the usefulness of open-vocabulary edges.
Their current effectiveness may be primarily limited by the reliability of automatic relation generation, making robust and consistent edge extraction a key direction for future work.
Beyond relation quality, this approach encounters scalability challenges, as larger scenes consist of more objects and edges between them, which may exceed the \ac{llm}'s token limit.
Further research could address these limitations by improving the handling of \acp{3dsg}, similar to SayPlan~\cite{rana2023sayplangroundinglargelanguage}, by identifying relevant subgraphs, or borrowing ideas from Retrieval Augmented Generation (RAG)~\cite{lewis2021retrievalaugmentedgenerationknowledgeintensivenlp}.
In addition, extending the evaluation to a broader range of \acp{llm} and \acp{vlm} and prompting techniques (\eg{} chain of thought prompting~\cite{wei2023chainofthoughtpromptingelicitsreasoning})  could further support the generalizability of the results.

Finally, our study shows that especially more capable models seem to benefit from explicit edges the most, with \ac{gpt_5} reaching above 99.5\% accuracy on some datasets, compared to the baseline accuracy of 73.0\%.

Taken together, these results suggest that providing modern \acp{llm} with explicit spatial structure from \acp{3dsg} enables more reliable object grounding and offers a concrete path toward robust robotic systems capable of natural language interaction in complex human environments.

%% file: Chapters/Acknowledgments.tex
\section{Acknowledgments}

The authors would like to acknowledge the use of Grammarly's generative AI tool~\cite{grammarly} and OpenAI's GPT4o~\cite{openai_models2025} for improving readability across all sections. Additionally, the authors acknowledge the role of OpenAI's GPT4o in generating documentation and improving code readability in the project's code base.